\documentclass[10pt, conference]{IEEEtran}

\IEEEoverridecommandlockouts

\usepackage[a4paper, left=54pt, right=37pt, top=122pt, bottom=54pt]{geometry}
\newgeometry{left=54pt, right=37pt, top=122pt, bottom=54pt}

\usepackage{cite}
\usepackage{amsmath, amssymb, amsfonts}
\usepackage{graphicx}
\usepackage{tikz}
\usepackage{textcomp}
\usepackage{multirow}

\usepackage{algpseudocode}
\usepackage{algorithm}

\usepackage{subcaption}
\usepackage{url}
\usepackage{lipsum}
\usepackage[utf8]{inputenc}

\begin{document}

\title{Visual Prompting for Robotic Manipulation with Annotation-Guided Pick-and-Place Using ACT\\
}

\author{\IEEEauthorblockN{1\textsuperscript{st} Muhammad A. Muttaqien\IEEEauthorrefmark{1}\IEEEauthorrefmark{2}}
\IEEEauthorblockA{\textit{Embodied AI Research Team} \\
\textit{National Institute of AIST}\\
Tokyo, Japan \\
muha.muttaqien@aist.go.jp}
\and
\IEEEauthorblockN{2\textsuperscript{nd} Tomohiro Motoda\IEEEauthorrefmark{1}}
\IEEEauthorblockA{\textit{Embodied AI Research Team} \\
\textit{National Institute of AIST}\\
Tokyo, Japan \\
tomohiro.motoda@aist.go.jp}
\and
\IEEEauthorblockN{3\textsuperscript{rd} Ryo Hanai}
\IEEEauthorblockA{\textit{Embodied AI Research Team} \\
\textit{National Institute of AIST}\\
Tokyo, Japan \\
ryo.hanai@aist.go.jp}
\and
\IEEEauthorblockN{4\textsuperscript{th} Yukiyasu Domae}
\IEEEauthorblockA{\textit{Embodied AI Research Team} \\
\textit{National Institute of AIST}\\
Tokyo, Japan \\
domae.yukiyasu@aist.go.jp}
}

\maketitle

\footnotetext[1]{These authors contributed equally to this work.}

\footnotetext[2]{Also affiliated with the Department of Computer Science, University of Tsukuba, Tsukuba, Japan.}

\begin{abstract}
Robotic pick-and-place tasks in convenience stores pose challenges due to dense object arrangements, occlusions, and variations in object properties such as color, shape, size, and texture. These factors complicate trajectory planning and grasping. This paper introduces a perception-action pipeline leveraging annotation-guided visual prompting, where bounding box annotations identify both pickable objects and placement locations, providing structured spatial guidance. Instead of traditional step-by-step planning, we employ Action Chunking with Transformers (ACT) as an imitation learning algorithm, enabling the robotic arm to predict chunked action sequences from human demonstrations. This facilitates smooth, adaptive, and data-driven pick-and-place operations. We evaluate our system based on success rate and visual analysis of grasping behavior, demonstrating improved grasp accuracy and adaptability in retail environments.
\end{abstract}

\begin{IEEEkeywords}
Visual Prompting, Imitation Learning, Object Selection, Robotic Manipulation
\end{IEEEkeywords}

\begin{figure*}[tbp]
  \centering
  \includegraphics[width=\linewidth]{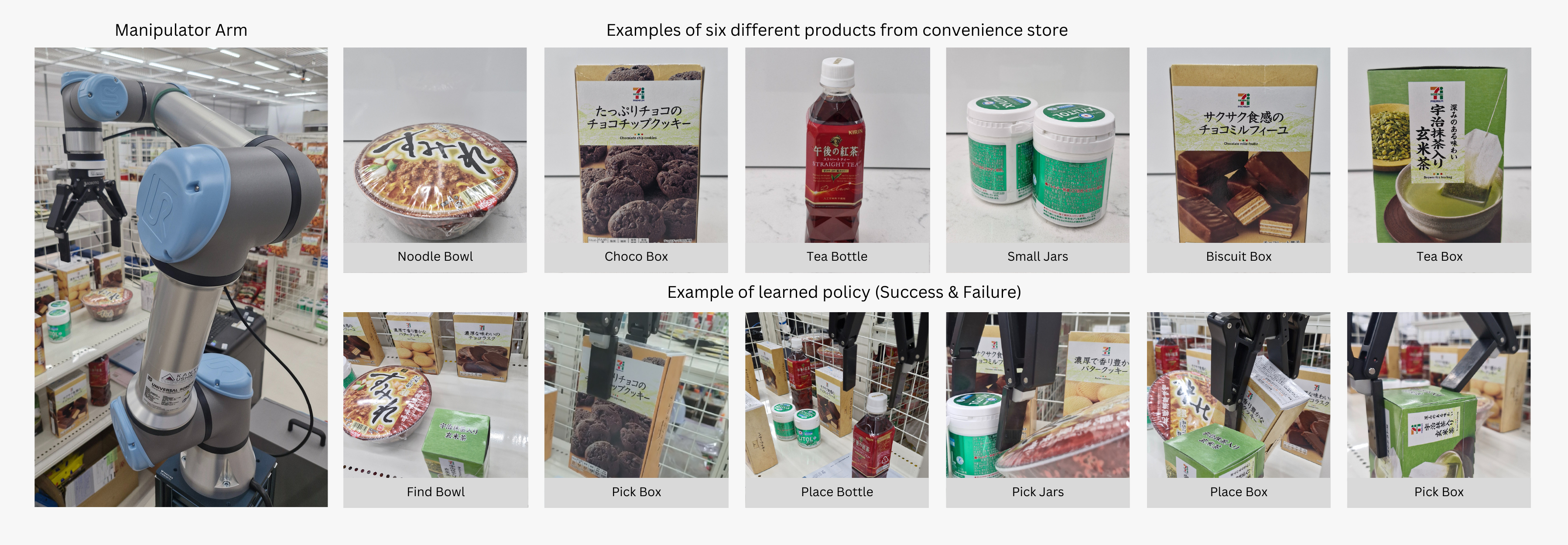}
  \caption{The perception-to-action system for robotic manipulation is trained using an imitation learning algorithm (ACT) and handles various products typically found in convenience stores across Japan. These products come in different shapes, colors, textures, and sizes, which influence the system's strategy for picking and placing items.}
  \label{fig:overview}
  \vspace{-2mm}
\end{figure*}

\section{Introduction}
Robotic pick-and-place tasks are essential in various industrial and retail applications, particularly in convenience stores where robots must handle a diverse range of products with different shapes, sizes, textures, and colors, as shown in Figure \ref{fig:overview}. However, real-world pick-and-place scenarios pose significant challenges due to dense object arrangements, frequent occlusions, and the need for precise grasping and placement. Traditional approaches often rely on predefined heuristics, structured environments, or explicit object segmentation, which lack adaptability when faced with novel items and dynamic layouts \cite{Saxena2006, Collet2011, Zhang2022}.

To address these challenges, this paper proposes a perception-action approach leveraging annotation-guided visual prompting for robotic manipulation. The system employs bounding box annotations to provide structured spatial guidance for both object selection and placement, reducing the reliance on exhaustive scene understanding. This method enables robots to interpret and act upon visual cues efficiently, making it highly suitable for retail environments where objects frequently change. Instead of relying on conventional step-by-step planning, we integrate Action Chunking with Transformers (ACT) as an imitation learning algorithm, allowing the robotic arm to predict chunked action sequences based on human demonstrations. This approach facilitates smoother, more adaptive pick-and-place operations by leveraging learned action patterns rather than predefined trajectories.

To systematically evaluate the effectiveness of annotation-guided visual prompting, we conduct experiments across three levels of task complexity:

\begin{itemize}
    \item \textbf{Simple Scenario}: Nine products of similar shape and size (boxes) are arranged in a 3x3 format, with one object annotated for picking.
    \item \textbf{Complex Scenario}: Nine diverse products (varying in shape, color, texture, and size) are arranged in a 3x3 format, with one object annotated for picking.
    \item \textbf{More Complex Scenario}: Nine diverse products are placed in varying positions, with one annotated for picking and another location annotated as the destination.
\end{itemize}

Through extensive evaluation, we demonstrate that our approach enhances grasp accuracy, and adaptability. The contributions of this work are summarized as follows: 

\begin{itemize}
    \item \textbf{Annotation-Guided Visual Prompting}: We introduce bounding box-based visual prompting to guide robotic pick-and-place manipulation, reducing perception complexity while ensuring task execution.
    \item \textbf{ACT-based System Implementation for Adaptive Pick-and-Place}: We apply ACT as an imitation learning algorithm, enabling the robotic arm to execute smooth, chunked action sequences instead of rigid, step-by-step planning.
    \item \textbf{Systematic Evaluation Across Progressive Task Complexity}: We present a structured experimental framework to analyze how visual prompting influences robotic performance across different levels of object diversity and scene complexity.
\end{itemize}

By combining visual prompting and action chunking, this study provides insights into how robots can efficiently interpret and act upon minimal yet informative annotations for real-world manipulation tasks. The proposed method represents a step forward in adaptive robotic systems capable of complex object variations with better autonomy and robustness.

\section{RELATED WORK}

Robotic pick-and-place tasks have been extensively studied in the domains of robotic perception, grasping, and motion planning. Traditional approaches often rely on explicit scene understanding, predefined grasp heuristics, and step-by-step planning. However, these methods struggle with real-world constraints such as dense object arrangements, occlusions, and diverse object properties. Recent advancements in visual prompting, imitation learning, and deep learning-based action prediction have opened new possibilities for more adaptive and efficient robotic manipulation. In this section, we discuss related works in vision-based robotic grasping, visual annotation for manipulation, imitation learning for action chunking, and task evaluation in complex environments.

Vision-based grasping methods have traditionally relied on depth sensors and model-based grasp synthesis to estimate object poses and predict stable grasp configurations \cite{Zhang2022}. While these methods are effective in structured environments, they struggle with dynamic and cluttered settings. Deep learning-based approaches such as DexGraspNet \cite{Wang2023} and Improved GQ-CNN \cite{Jaskowski2020} have demonstrated how convolutional neural networks (CNNs) can learn robust grasping strategies from large-scale datasets. However, these methods often require extensive training and do not generalize well to novel object arrangements without retraining. Unlike these approaches, our method introduces annotation-guided visual prompting, where bounding boxes serve as lightweight yet effective spatial cues for guiding robotic grasping and placement. This reduces the need for exhaustive scene parsing while maintaining high grasp accuracy.

The use of visual annotations, such as bounding boxes, segmentation masks, and keypoints, has gained attention as a means of guiding robotic manipulation. Previous works have explored human-in-the-loop annotation systems, where human operators provide bounding box guidance for object selection \cite{Leeper2012}. Additionally, visual affordance-based methods \cite{Zeng2018} have been used to predict graspable regions based on object features. More recently, CLIPort \cite{Shridhar2021}, PIVOT \cite{Nasiriany2024}, and MOKA \cite{Fang2024} have demonstrated how language-conditioned visual prompting can improve object manipulation by combining textual and visual cues. However, these approaches often rely on complex scene parsing or multimodal fusion, making them computationally expensive. Our research simplifies visual prompting by demonstrating that bounding boxes alone can effectively guide robotic pick-and-place actions.

Imitation learning has been widely used in robotic manipulation, enabling robots to learn from expert demonstrations rather than relying solely on reinforcement learning. Classical methods such as Behavior Cloning (BC) and Inverse Reinforcement Learning (IRL) have been applied to various robotic tasks. However, these methods suffer from error accumulation, where small mistakes compound over time, leading to failure in long-horizon tasks. To address this, Action Chunking with Transformers (ACT) \cite{Zhao2023} introduces a hierarchical action representation that enables robots to predict sequence-level actions rather than individual step-by-step motions. Our research leverages ACT as an imitation learning algorithm, showing that chunked action sequences improve grasp stability, and placement accuracy when guided by bounding box-based visual prompts.

Prior research has explored various strategies for evaluating robotic grasping and object placement. Probabilistic models for grasp transfer \cite{Kopicki2016} have been used to generalize learned grasping strategies across different object categories, while reinforcement learning-based grasp learning \cite{Levine2018} has demonstrated the ability to train robots in large-scale real-world environments. More recently, DexNet \cite{Mahler2017DexNet} has shown that data-driven grasp synthesis can improve grasp robustness across diverse objects. However, many of these studies focus primarily on grasp execution without explicitly considering placement accuracy and task adaptability. Our research extends prior work by systematically evaluating both grasping and placement performance under progressively complex scenarios. By analyzing how object diversity (shape, size, texture) and annotation positioning impact robotic manipulation capabilities.

\section{ROBOT SETUP}

The robotic manipulation system used in this study features a Universal Robots UR5e arm equipped with a gripper for picking and placing objects. The setup includes a vision system that utilizes a camera to capture real-time images of the workspace for object detection. Both real-world and simulation experiments were conducted using the RoboManipBaselines framework~\cite{RoboManipBaselines_GitHub2024}, a robot learning software platform. This section provides a detailed overview of the system’s hardware, software, and component integration.

\subsection{Robotic Arm}
\begin{figure}[h!]
    \centering
    \includegraphics[width=\linewidth]{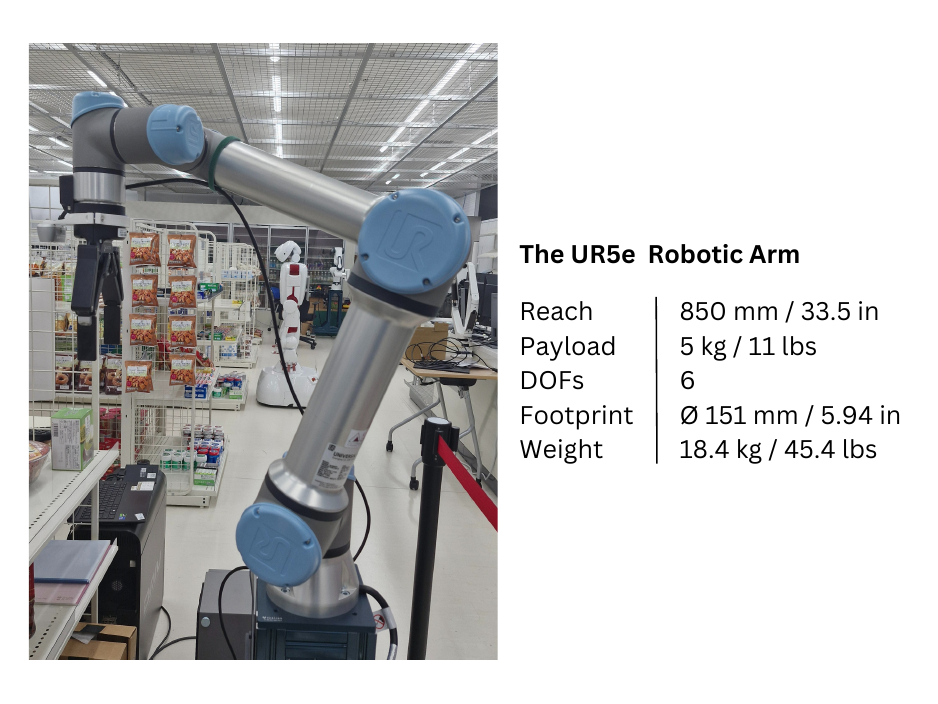}
    \caption{A snapshot of the manipulator arm (UR5e), a lightweight and versatile cobot, along with its specifications, is demonstrated in this study.}
    \label{fig:specification}
\end{figure}

The UR5e is a lightweight, versatile collaborative robot (cobot) designed for industrial and research applications. As the larger of Universal Robots' two light payload cobots, it offers a balance of reach, payload, and precision, making it ideal for precise tasks like object grasping, placement, and assembly. This 6-DOF robotic arm delivers high precision and smooth movements, essential for precise operations, as shown in Figure \ref{fig:specification}. The UR5e can be controlled via ROS and integrated into the system through a custom driver that allows real-time communication between the robot and the control software. 

\subsection{Gripper}
The robotic arm is paired with a Robotiq 2-Finger Gripper, known for its exceptional adaptability and versatility. Designed for handling objects of various shapes and sizes, it supports both encompassing and pinch grips, ensuring precise and powerful handling across different applications. The gripper integrates seamlessly with the ROS \cite{Quigley2009} interface, enabling real-time control based on task requirements. With a payload capacity of 2.5 kg (5.5 lb) and a tooling weight of 1 kg (2 lb), the gripper features a stroke length of 140 mm (5.5 in.) and a grip force ranging from 10 to 125 N (2 to 25 lbf). The closing speed varies between 30 to 250 mm/s (1.2 to 9.8 in./s), and it provides control over essential parameters such as object detection, position, speed, and force. 

\subsection{Vision System}
The system employs two Intel RealSense cameras for visual perception, providing depth and RGB data to enhance robotic manipulation accuracy. A RealSense D435i is mounted on the robot’s hand, enabling close-range perception and dynamic viewpoint adjustments, while a RealSense D415 is positioned at the front of the tabletop, offering a stable, wide-angle perspective of the workspace. 

The RealSense D435i, mounted on the hand, offers an RGB frame resolution of 1920 × 1080, a Depth Field of View (FOV) of 87° × 58°, and connects via USB 3.0 for high-speed data transfer. The RealSense D415, positioned on the tabletop, also provides an RGB frame resolution of 1920 × 1080, with a Depth Field of View (FOV) of 65° × 40°, and similarly connects via USB 3.0. 

The captured images (without depth data) from both cameras are directly fed into ACT’s CNN network, allowing for real-time scene understanding, grasp prediction, and precise object interaction without explicit bounding box annotations.

\subsection{Workspace Configuration}
The robot operates in a controlled workspace designed to simulate a convenience store setting. The workspace includes a tabletop with a variety of objects placed in different configurations. These objects are arranged to test the system’s ability to handle and diverse product types, mimicking the real-world challenges faced in retail environments. The workspace is equipped with reference markers for calibration and alignment of the camera and robot arm, ensuring accurate task execution.

\section{ACTION CHUNKING WITH TRANSFORMERS}

Action Chunking with Transformers (ACT) is a novel imitation learning algorithm designed to enable robots to learn complex action sequences from human demonstrations. The key motivation behind ACT is to improve the efficiency, adaptability, and robustness of robotic manipulation in dynamic and complex environments. In traditional robotic control, long-horizon tasks, such as pick-and-place operations, require precise sequencing of actions that can easily become error-prone when relying on small, incremental steps. By chunking actions into higher-level sequences, ACT allows the robot to predict longer, more coherent sequences of actions, improving the efficiency of task execution.

\begin{figure*}[tbp]
\vspace{-20mm}
  \centering
  \includegraphics[width=\linewidth]{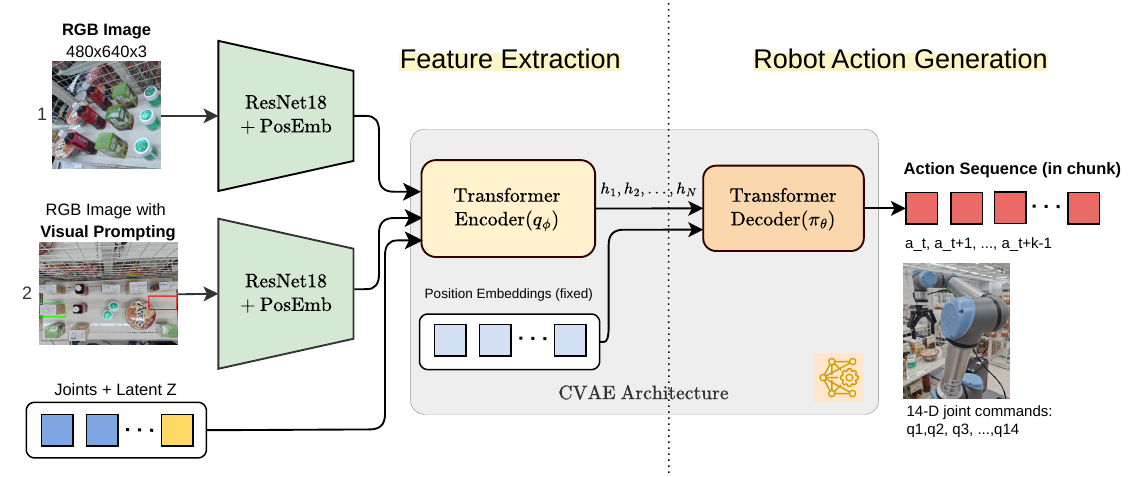}
  \caption{Architecture of the proposed ACT framework with visual prompting. The model receives both the original RGB image and a visually prompted RGB image, where task-relevant objects and target locations are highlighted using bounding boxes (green for picking and red for placing). Features extracted from both image streams are fused with robot joint states and a latent style variable (z) obtained from a CVAE encoder. These representations are processed by a transformer encoder, while a transformer decoder generates a chunk of future robot actions represented as 14-dimensional joint commands.}
  \label{fig:diagram}
  \vspace{-2mm}
\end{figure*}

ACT employs a Transformer-based architecture to model action sequences. The Transformer, known for its ability to handle sequential data, serves as the backbone of the ACT algorithm, enabling the robot to learn the temporal dependencies and relationships between actions. Unlike traditional sequential models such as Long Short-term Memory (LSTMs) \cite{Hochreiter1997}, Transformers \cite{Vaswani2017} are better equipped to capture long-range dependencies, which is crucial for complex tasks involving multiple steps. A diagram illustrating our overall system architecture based on ACT is shown in Figure~\ref{fig:diagram}. Basically, the structure of ACT can be broken down into the following components:

\subsection{Action Chunking}
Instead of predicting each individual action, ACT groups related actions into ``chunks.'' These chunks represent coherent segments of the task that can be executed together, allowing the robot to perform multiple subtasks in parallel or with fewer intermediate steps. For example, a pick-and-place task might be chunked into the ``picking'' and ``placing'' actions, with each chunk corresponding to a higher-level goal.

\subsection{Transformer Encoder}
The Transformer encoder processes the sequence of demonstration data, which includes human-provided action sequences along with sensory inputs (such as visual data or force feedback). The encoder learns to encode these sequences into a rich representation of the task, capturing both the temporal relationships between actions and the spatial relationships between objects.

\subsection{Action Prediction}
Once the action sequences have been chunked and encoded, ACT uses the Transformer decoder to predict the next chunked action sequence. This prediction is based on the encoded information and the robot’s current state, such as the position of the object, its grasp status, and the target placement location.

\subsection{Output Generation}
The predicted action sequences are then converted into robot commands, which drive the robotic arm and gripper to perform the desired actions. These commands are executed in real-time, with the robot continuously adjusting its behavior based on feedback from the environment.

ACT relies on human demonstrations to learn action sequences. These demonstrations provide the robot with high-level examples of how to perform the task in a variety of scenarios. The learning process is structured as a form of imitation learning, where the robot observes the actions performed by a human and learns to replicate them. The training process for ACT is outlined in Algorithm \ref{alg:act_train}.

\begin{algorithm}
\caption{ACT Training Algorithm with Visual Prompting}
\label{alg:act_train}
\begin{algorithmic}[1]
\State \textbf{Given:} Demo dataset $\mathcal{D}$, chunk size $k$, weight $\beta$.
\State Let $a_t$, $o_t$ represent the action and observation at timestep $t$, $\bar{o}_t$ represent $o_t$ without image observations.
\State Let $I_t$ be the RGB image at timestep $t$, and $I^{\text{vp}}_t$ be the same image with visual prompts (e.g., bounding boxes).
\State Initialize encoder $q_{\phi}(z|a_{t:t+k}, \bar{o}_t)$.
\State Initialize decoder $\pi_{\theta}(\hat{a}_{t:t+k}|I^{\text{vp}}_t, o_t, z)$.
\For{$n = 1, 2, \dots$}
    \State Sample $o_t, a_{t:t+k}, I_t$ from $\mathcal{D}$.
    \State Generate $I^{\text{vp}}_t$ by adding visual prompts to $I_t$.
    \State Sample $z$ from $q_{\phi}(z|a_{t:t+k}, \bar{o}_t)$.
    \State Predict $\hat{a}_{t:t+k}$ from $\pi_{\theta}(\hat{a}_{t:t+k}|I^{\text{vp}}_t, o_t, z)$.
    \State $L_{\text{reconst}} = \text{MSE}(\hat{a}_{t:t+k}, a_{t:t+k})$.
    \State $L_{\text{reg}} = D_{\text{KL}}(q_{\phi}(z|a_{t:t+k}, \bar{o}_t) \parallel \mathcal{N}(0, I))$.
    \State Update $\theta, \phi$ with ADAM and $\mathcal{L} = L_{\text{reconst}} + \beta L_{\text{reg}}$.
\EndFor
\end{algorithmic}
\end{algorithm}

\section{EXPERIMENTS}

\subsection{Tasks} 

In this study, six different products commonly found in convenience stores in Japan, are used to evaluate the performance of the robotic system. These products represent a diverse range of packaging types, sizes, shapes, and textures, and each requires the robot to perform both pick-and-place tasks. The variety of objects evaluate the system’s adaptability and effectiveness in handling real-world products in a retail environment.

The \textbf{noodle bowl} is a round, medium-sized container with a silver body and a colorful lid featuring Japanese text. It has a smooth plastic exterior with a foil-sealed top, presenting challenges in grasping due to its round shape and foil lid. The \textbf{choco box}, a small to medium rectangular box with a beige background and dark brown cookie images, is lightweight and easy to handle due to its smooth cardboard texture. Similarly, the \textbf{biscuit box}, also rectangular in shape and small to medium in size, has a light brown design with dark brown biscuit images. It shares the smooth cardboard surface of the choco box, requiring the robot to handle it carefully, considering its relatively lightweight nature. The \textbf{tea box}, a small rectangular package with a green design and an image of a steaming cup of green tea, stands out with its compact size and smooth cardboard texture. The robot is tasked with picking up and placing this product, navigating the challenges of its small dimensions and distinctive green color.

The \textbf{tea bottle}, a tall, slim plastic bottle containing reddish-brown liquid, features slight ridges for grip and is sealed with a white plastic cap. The slender shape and ridges require precise handling for secure grasping, making it more challenging for the robot to manipulate. Finally, the \textbf{small jars}, small cylindrical containers with white screw-top lids, are typically used for supplements or cosmetics. These jars have a smooth plastic texture and are easy to store and carry, but their cylindrical shape presents unique challenges in ensuring proper handling and placement.

These six products, with distinct packaging, material, and visual appeal, serve to test the system's grasping accuracy, adaptability to diverse shapes, and effectiveness in a real-world retail environment. The tasks focus on evaluating how well the system can adapt to variations in packaging, handle various object types, and execute pick-and-place actions using annotation-guided visual prompting and ACT.

\subsection{Data Collection} 

To collect the demonstration data for training the ACT model, we used a 3D space mouse for teleoperation. This setup allowed a human operator to control the robotic arm in a precise and intuitive manner, providing demonstrations for each pick-and-place task. Each demonstration episode lasted between 21-30 seconds, depending on the complexity of the grasping and trajectory required to reach the placement destination. The total number of demonstrations recorded for each product was 10, which provided a sufficient sample size for training while maintaining manageable data collection times.

The total duration of data collected for each product ranged from 3 to 5 minutes. This was based on the 10 demonstrations, with each demonstration lasting between 21-30 seconds. For the 3x3 setup, the grand total of demonstration data collected was between 27 and 45 minutes, depending on the specific products involved.

It is important to note that all human demonstrations are inherently stochastic, even though a single person was responsible for performing all of the demonstrations. Variability in human execution is natural, and the different characteristics of each product, such as size, shape, and packaging, led to variations in the way each object was grasped. Additionally, the placement location for each product introduced further variability in the robot's trajectory to reach the destination.

\begin{figure}[h!]
    \centering
    \includegraphics[width=\linewidth]{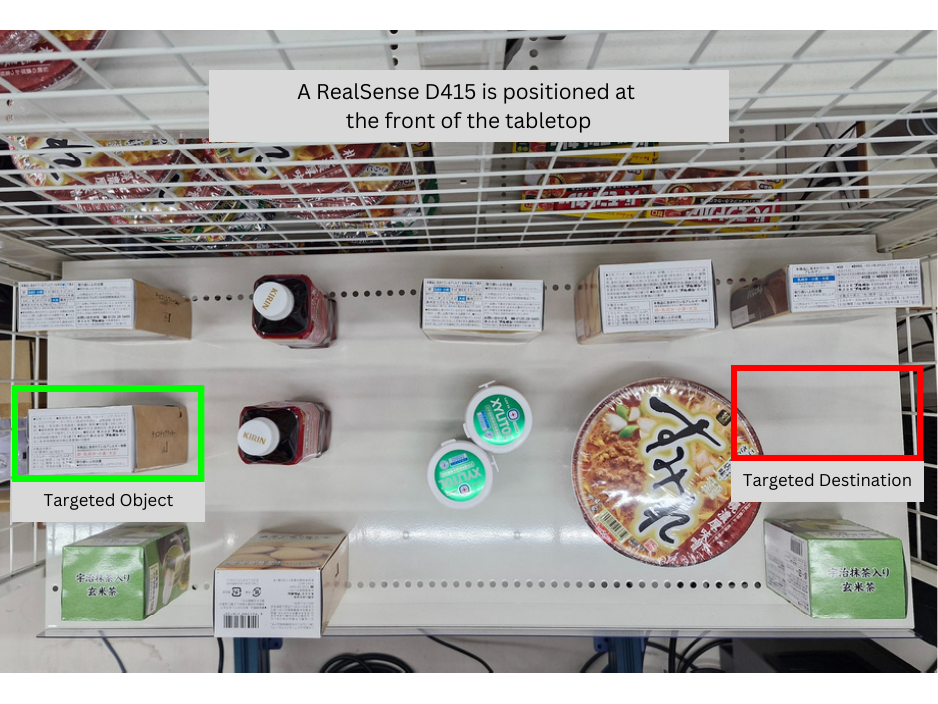}
    \caption{The bird’s-eye view of the experimental setup shows a variety of products placed in various locations. The green bounding box indicates the targeted object, while the red one indicates the targeted destination.}
    \label{fig:attention}
    \vspace{-4mm}
\end{figure}

\subsection{ACT Hyper-parameters Tuning}

To optimize the performance of the Action Chunking with Transformers (ACT) model, we conducted hyper-parameter tuning using a simulation environment built on MuJoCo \cite{Todorov2012}. The model was implemented using PyTorch \cite{Paszke2019}, which provided flexibility and ease of implementation for our deep learning tasks. The following three hyper-parameters were considered for tuning: batch size, chunk size, and epoch count. These hyperparameters were adjusted to assess their impact on model performance and training efficiency.

\begin{itemize}
    \item \textbf{Batch Size:} We tested a range of batch sizes, from 8 to 2. Smaller batch sizes introduce more noise, which can help the model avoid overfitting to the training data.
    
    \item \textbf{Chunk Size:} The chunk size was varied between 100 and 140. A chunk size of 100 was selected as the final choice, balancing flexibility and efficiency. Larger chunks reduce learning flexibility, while smaller ones enhance adaptability.
    
    \item \textbf{Epoch Count:} We tested different epoch values ranging from 1000 to 1600. The final selection was 1600 epochs, which provided sufficient time for the model to learn and improve, with room for further enhancement.
\end{itemize}

Based on the results of these experiments, batch size and epoch count emerged as the two most critical hyperparameters for improving the performance, specifically the success rate of the trained model. It was observed that increasing the number of epochs provided the model with more data to train on, potentially resolving issues where the gripper failed to grasp the item. The longer training duration and smaller batch size were expected to enhance the model’s ability to generalize and handle a variety of object types more effectively.

\subsection{Experiment Results}

\begin{figure*}[tbp]
  \centering
  \includegraphics[width=\linewidth]{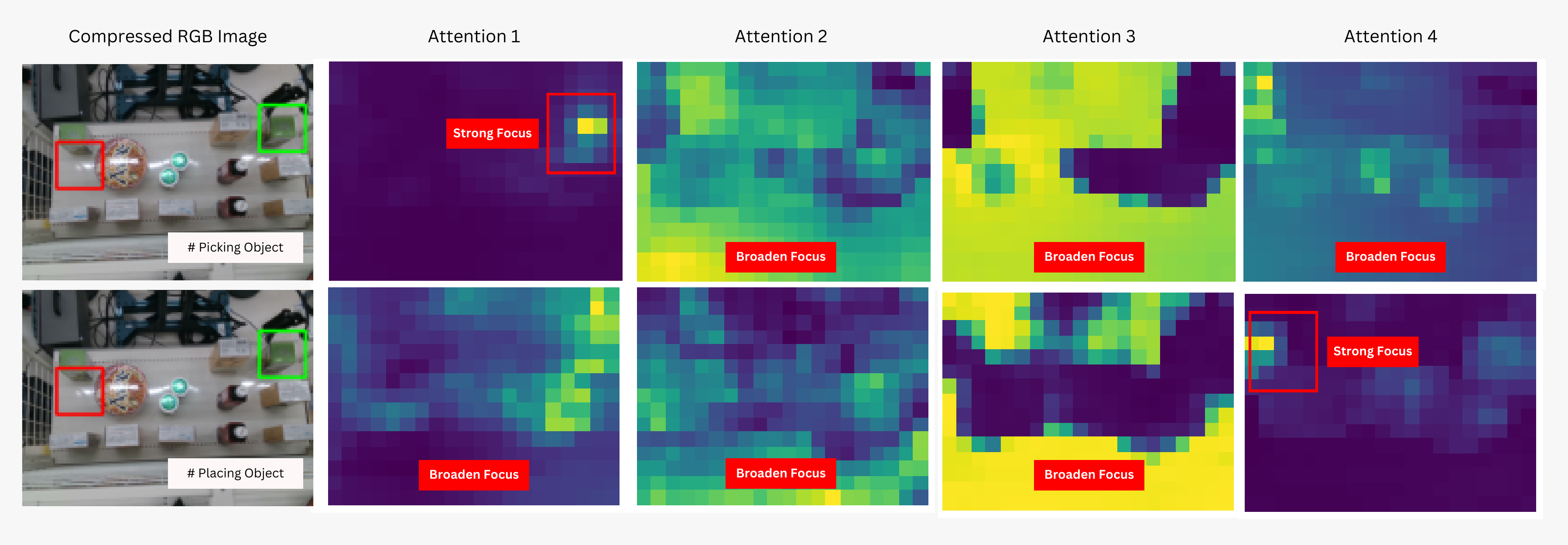}
  \caption{Attention heatmaps are used to assess where ACT focuses when processing bounding box prompts for performing picking and placing tasks. These heatmaps are generated based on different transformer's encoder layers.}
  \label{fig:attention}
  \vspace{-2mm}
\end{figure*}

The effectiveness of annotation-guided visual prompting was evaluated across three levels of task complexity: simple, complex, and more complex scenarios. These scenarios used five products (\textbf{choco box}, \textbf{biscuit box}, \textbf{tea box}, \textbf{tea bottle}, and \textbf{small jars}) to assess how well bounding box annotations guide the robotic arm in different environments. Each product was tested with 10 trials to ensure consistent evaluation. The \textbf{noodle bowl} was included in the setup to increase the complexity of the scene but was not considered as a product to be grasped and placed due to its graspability issues.

In the \textbf{simple scenario}, where nine products of similar shape and size (boxes) were arranged in a 3x3 format with one object annotated for picking, the robotic system achieved a 90\% success rate. The bounding box annotations effectively guided the robotic arm, as the uniformity of the products and their easy accessibility led to minimal challenges. This high success rate indicated that the system could reliably follow the bounding box cues in straightforward environments. In the \textbf{complex scenario}, where nine diverse products were arranged in a 3x3 format with one object annotated for picking, the success rate initially dropped to 70\%. This drop can be attributed to the increased variation in object characteristics, such as differing shapes, colors, and textures. Despite the added complexity, the bounding box annotations still guided the robotic arm effectively, though the increased variations led to some performance degradation. However, after increasing the demonstration data by 20\% for the failure cases, the system’s performance improved significantly, achieving a 100\% success rate in this scenario. This enhancement demonstrates the importance of more diverse training data in addressing the challenges posed by varied object types environments. In the \textbf{more complex scenario}, where nine diverse products were placed in varying positions with one annotated for picking and another for placement, the success rate initially decreased to 70\%. The robot had to navigate around other objects to reach its destination, introducing challenges such as occlusions and more complex trajectories. Since the task proved to be more difficult, we collected double the amount of human-annotated data for each product (as shown in Table \ref{tab:success_rate}).

\begin{table}[h!]
\centering
\caption{Success Rate Across Different Scenarios: Simple Scenario (Scenario 1), Complex Scenario (Scenario 2), and More Complex Scenario (Scenario 3), and Object Categories.}
\begin{tabular}{|p{2.25cm}|l|c|}
\hline
\textbf{Scenario} & \textbf{Object Category} & \textbf{Success Rate} \\ [1.5ex] \hline
\multirow{1}{*}{Scenario 1} & Rigid 1 (Choco Box, Biscuit Box) & 90\% \\ [0.75ex] \hline
\multirow{5}{*}{Scenario 2} & Rigid 1 (Choco Box, Biscuit Box) & 70\% \\ [0.75ex]
                                 & Rigid 2 (Smaller Tea Box) & 0\% \\ [0.75ex]
                                 & Reflective (Tea Bottle) & 100\% \\ [0.75ex]
                                 & Slippery (Small Jars) & 0\% \\ [0.75ex]
                                 & Flexible (Noodle Bowl) & Excluded \\ [0.75ex] \hline
\multirow{5}{*}{\shortstack{Scenario 2 \\ (More Data)}} & Rigid 1 (Choco Box, Biscuit Box) & 90\% \\ [0.75ex]
                                 & Rigid 2 (Smaller Tea Box) & 100\% \\ [0.75ex]
                                 & Reflective (Tea Bottle) & 100\% \\ [0.75ex]
                                 & Slippery (Small Jars) & 100\% \\ [0.75ex]
                                 & Flexible (Noodle Bowl) & Excluded \\ [0.75ex] \hline
\multirow{5}{*}{Scenario 3} & Rigid 1 (Choco Box, Biscuit Box) & 90\% \\ [0.75ex]
                                 & Rigid 2 (Smaller Tea Box) & 90\% \\ [0.75ex]
                                 & Reflective (Tea Bottle) & 80\% \\ [0.75ex]
                                 & Slippery (Small Jars) & 80\% \\ [0.75ex]
                                 & Flexible (Noodle Bowl) & Excluded \\ [0.75ex] \hline
\end{tabular}
\label{tab:success_rate}
\vspace{-2mm}
\end{table}

To explore how ACT allocates its focus when processing bounding box prompts, attention heatmaps were generated for each scenario. In simpler settings, ACT primarily focused on the center of the bounding boxes. As shown in Image~\ref{fig:attention}, the attention quickly shifts from the location of the picking object to the location of the placing object. This shift in focus indicates that ACT understands the need to navigate between these locations to accomplish the pick-and-place tasks based on the visual prompting provided. This adaptive focus behavior highlights how ACT adjusts its attention strategy in response to increased environmental complexity.

The system's performance was also analyzed across different object categories to determine whether the bounding box prompt worked equally well for objects of varying shapes, colors, textures, and sizes. The comparative analysis revealed that bounding box prompts worked particularly well for rigid objects like the \textbf{choco box} and \textbf{biscuit box}, which had a high success rate of 90\%. However, objects with reflective surfaces, such as the \textbf{tea bottle}, achieved a lower success rate (before adding more data), while soft or flexible objects, like the \textbf{noodle bowl}, were excluded from the evaluation due to their graspability issues. These results suggest that while bounding box annotations guide the robot effectively.

Finally, visualizations of grasping errors highlighted how certain object properties posed more challenges for the system. Reflective surfaces caused misalignment issues, leading to a moderate failure rate (before adding more data) due to difficulty in accurately detecting the object’s edges. These errors emphasize that certain object characteristics, such as reflectivity and softness, complicate the effectiveness of visual prompting.

\begin{figure}[h!]
    \centering
    \begin{subfigure}[b]{0.30\linewidth}
        \centering
        \includegraphics[width=\linewidth]{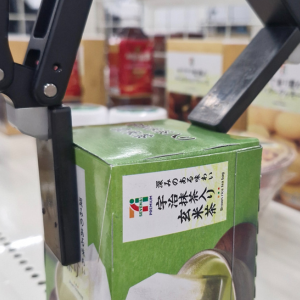}
        \caption{Finger Misalignment}
    \end{subfigure}
    \hfill
    \begin{subfigure}[b]{0.30\linewidth}
        \centering
        \includegraphics[width=\linewidth]{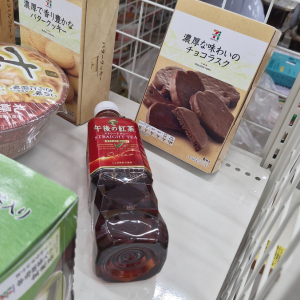}
        \caption{Placement Failure}
    \end{subfigure}
    \hfill
    \begin{subfigure}[b]{0.30\linewidth}
        \centering
        \includegraphics[width=\linewidth]{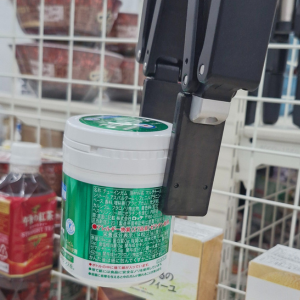}
        \caption{Too Slippery Surface}
    \end{subfigure}
    
    \vspace{4mm} 
    
    \begin{subfigure}[b]{0.30\linewidth}
        \centering
        \includegraphics[width=\linewidth]{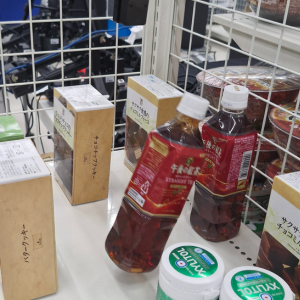}
        \caption{Weak Grasp Power}
    \end{subfigure}
    \hfill
    \begin{subfigure}[b]{0.30\linewidth}
        \centering
        \includegraphics[width=\linewidth]{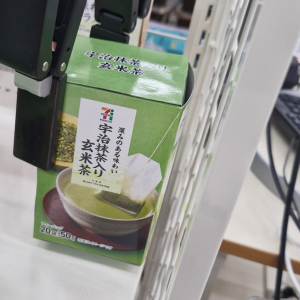}
        \caption{Placement Misalignment}
    \end{subfigure}
    \hfill
    \begin{subfigure}[b]{0.30\linewidth}
        \centering
        \includegraphics[width=\linewidth]{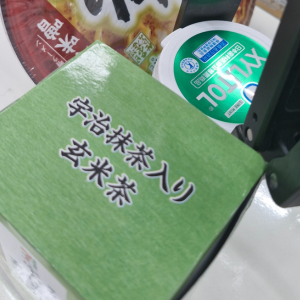}
        \caption{Narrow Destination}
    \end{subfigure}

    \caption{Different products to pick and different locations to place can potentially lead to failures if not handled with the right strategy.}
    \label{fig:failure}
    \vspace{-4mm}
\end{figure}

\subsection{Failure Analysis}

Figure \ref{fig:failure} illustrates that the system encountered several failure modes when performing pick-and-place tasks, particularly during the early stages of training when it was underlearned. Finger misalignment occurred when the gripper's fingers failed to align properly with the object, preventing a secure grasp. Additionally, weak grasp power caused items to slip or fall during the trajectory, especially when handling objects with slippery surfaces. In some cases, placement failure occurred when the object was dropped during movement, while placement misalignment happened when the object was placed incorrectly due to slight miscalculations.

A narrow destination also proved challenging, making it difficult for the robot to place objects accurately in tight spaces. These issues were most prominent when the system had not yet fully learned to handle these tasks, highlighting areas for improvement. Specifically, the failures point to the need for enhanced gripper precision, strength, and overall task adaptability as the system continues to learn and refine its performance.

\section{CONCLUSION}

In this paper, we have presented a novel approach to robotic pick-and-place tasks using annotation-guided visual prompting and Action Chunking with Transformers (ACT). By leveraging bounding box annotations, we reduce the complexity of scene understanding while maintaining high grasp accuracy and efficient task execution. The integration of ACT enables the robot to predict chunked action sequences, improving task adaptability and robustness in dynamic retail environments. Through systematic evaluation across varying task complexities, we demonstrate that our approach enhances the robot’s grasp stability, and placement accuracy. This work represents a step forward in creating more adaptive, data-driven robotic systems capable of performing real-world tasks.

However, this study remains data-demanding, as the effectiveness of the model relies heavily on diverse and high-quality human demonstration data. To address this limitation, future work will focus on developing a data-augmentation process to create human-like data artificially, helping to expand the dataset without requiring extensive manual demonstrations.

\section*{Acknowledgment}

We would like to express our sincere gratitude to the National Institute of Advanced Industrial Science and Technology (AIST) for their invaluable support and resources that made this research possible. Their contribution was essential in the successful completion of this research work.

\end{document}